\documentclass[journal]{IEEEtran}
\usepackage{booktabs}
\usepackage{tabularx}
\usepackage{amsmath,amssymb}
\usepackage{graphicx}
\usepackage{epstopdf}
\usepackage{float}
\usepackage{multirow}
\usepackage{cite}
\usepackage{booktabs}
\usepackage{subfigure}
\usepackage{xcolor}
\usepackage{amsfonts}
\usepackage{mathrsfs}
\usepackage{pifont}
\usepackage{stfloats}
\usepackage{color}
\usepackage{booktabs}
\usepackage{caption}
\usepackage{algorithm}
\usepackage{caption}
\usepackage{algorithmicx}
\usepackage{algpseudocode}
\usepackage{soul}
\renewcommand{\t}{^\top}

\IEEEoverridecommandlockouts                              
\newcommand{\m}{{\rm m}}
\newcommand{\eproof}{\hfill\rule{2mm}{2mm}}

\newcommand{\bstate}{\begin{state} }
	\newcommand{\estate}{ \hfill  \rule{1mm}{2mm}\end{state}}

\newcommand{\bass}{\begin{ass} }
	\newcommand{\eass}{\hfill\rule{1mm}{2mm}\end{ass}}
\newcommand{\bpro}{\begin{property}}
	\newcommand{\epro}{\hfill\rule{1mm}{2mm}\end{property}}

\newcommand{\brem}{ \begin{remark}  }
	\newcommand{\erem}{\hfill \rule{1mm}{2mm}
\end{remark} }
\newcommand{\bthm}{\begin{theorem}  }
	\newcommand{\ethm}{ \hfill  \rule{1mm}{2mm}
\end{theorem} }
\newcommand{\blem}{\begin{lemma}  }
	\newcommand{\elem}{ \hfill \rule{1mm}{2mm}
\end{lemma} }
\newcommand{\bcorollary}{\begin{corollary}  }
	\newcommand{\ecorollary}{  \hfill \rule{1mm}{2mm}
\end{corollary} }
\newcommand{\bdefn}{\begin{definition}}
	\newcommand{\edefn}{  \hfill \rule{1mm}{2mm}
\end{definition} }
\newcommand{\bproposition}{\begin{proposition} }
	\newcommand{\eproposition}{\hfill \rule{1mm}{2mm}
\end{proposition} }
\newcommand{\bexample}{\begin{example} \rm}
	\newcommand{\eexample}{ \hfill \rule{1mm}{2mm}
\end{example} }
\newcommand{\proofnow}{\noindent{\bf Proof: }}

\newtheorem{theorem}{\bf Theorem}[section]
\newtheorem{ass}{\bf Assumption}[section]
\newtheorem{lemma}{\bf Lemma}[section]
\newtheorem{definition}{\bf Definition}[section]
\newtheorem{remark}{\bf Remark}[section]
\newtheorem{corollary}{\bf Corollary}[section]
\newtheorem{proposition}{\bf Proposition}[section]
\newtheorem{example}{\bf Example}[section]

\newtheorem{property}{\bf Property}[section]

\newcommand{\sint}{\textstyle{\int}}

\usepackage{fancyhdr}
\allowdisplaybreaks[4]
\usepackage{bm} 
\begin{document}
	\title{Disturbance Estimation of Legged Robots: Predefined Convergence via Dynamic Gains}

	\author{Bolin Li,  Peiyuan Cai, Gewei Zuo, Lijun Zhu and Han Ding
		\thanks{ B. Li, P. Cai, G. Zuo, and L. Zhu are with School of Artificial Intelligence and Automation, Huazhong University of Science and Technology, Wuhan 430072, China. L. Zhu is also with Key Laboratory of Imaging Processing and Intelligence Control. H. Ding is with the State Key Laboratory of Intelligent Manufacturing Equipment and Technology, Huazhong University of Science and Technology (Emails:   bolin\_li@hust.edu.cn; cpy\_3566@163.com; gwzuo@hust.edu.cn; ljzhu@hust.edu.cn; dinghan@mail.hust.edu.cn).
			
	}}

	\markboth{}{ \MakeLowercase{\textit{et al.}}: }
	
	\maketitle
	
	\begin{abstract}
In this study, we address the challenge of disturbance estimation in legged robots by introducing a novel continuous-time online feedback-based disturbance observer that leverages measurable variables. The distinct feature of our observer is the integration of dynamic gains and comparison functions, which guarantees predefined convergence of the disturbance estimation error, including ultimately uniformly bounded, asymptotic, and exponential convergence, among various types. The properties of dynamic gains and the sufficient conditions for comparison functions are detailed to guide engineers in designing desired convergence behaviors. Notably, the observer functions effectively without the need for upper bound information of the disturbance or its derivative, enhancing its engineering applicability. An experimental example corroborates the theoretical advancements achieved.
	\end{abstract}
	
	\section{Introduction }
Disturbances are ubiquitous in legged robots and can significantly impair their locomotion. Consequently, disturbance rejection is a central goal in the design of locomotion algorithms for legged robots. When disturbances are measurable, it is well established that a feedforward strategy can mitigate or even nullify their effects. However, in many cases, external disturbances cannot be directly measured or the measurement equipment is prohibitively expensive. An intuitive solution to this challenge is to estimate the disturbance, or its effects, from measurable variables, and subsequently implement control actions based on these estimates to compensate for the disturbances' impact. 
	
Numerous disturbance observers have been proposed and implemented for robots, including the generalized momentum-based observer \cite{evangelisti2024data05}, joint velocity-based observer \cite{haddadin2017robot06}, nonlinear disturbance observer \cite{mohammadi2018hybrid07,kang2023external08}, disturbance Kalman filter \cite{hu2017contact09,mitsantisuk2011kalman10}, and extended state observer \cite{ren2019extended11, rsetam2019cascaded12}. Among these, the nonlinear disturbance observer and the extended state observer \cite{2025arXiv250210761L} have been the most extensively studied and applied. The nonlinear disturbance observer was introduced to address the limitations of linear disturbance observers, which rely on linear system techniques. On the other hand, the extended state observer offers an alternative practical control approach to the traditional PID methodology. It is generally regarded as a fundamental component of active disturbance rejection control, designed to estimate lumped disturbances arising from both unknown uncertainties and external disturbances.
	
The aforementioned disturbance observers typically employ constant feedback gains. The primary advantage of this design lies in its ease of application and the straightforward guarantee of stability. However, determining the appropriate value of the feedback gain presents several challenges. While a larger feedback gain can ensure the disturbance estimation error converges accurately, it may result in significant overshooting during the initial phase of operation. When these estimates are feedforward into the controller, they can cause the robot to execute excessive movements. Conversely, a smaller feedback gain effectively mitigates the issue of overshooting but compromises observation accuracy. To address these limitations, dynamic gains were introduced in \cite{krishnamurthy2020dynamic13, song2016adaptive15, song2017time14}, offering different types of dynamic gain structures tailored to achieve varying control performance objectives.

In this study, we investigate disturbance observers for legged robots, distinguishing our design by incorporating dynamic feedback gains coupled with comparison functions, as opposed to the conventional constant feedback gains. This approach enables the observer gains to grow at an optimal rate from a small initial value, effectively addressing overshoot and disturbance estimation error convergence issues. Consequently, when the disturbance observer are feedforward into the controller, they prevent excessive robot movements while significantly enhancing control performance. However, incorporating dynamic gains introduces new design challenges, particularly regarding the growth patterns of the dynamic gains and their coupling with the comparison functions. On one hand, we aim to explore dynamic gain types that promote diverse convergence behaviors of the disturbance estimation error, beyond a single convergence behavior. On the other hand, the coupling method between dynamic gains and comparison functions must exhibit broad adaptability to enhance practical applicability and scalability. The main innovations and contributions of our study are as follows: 
	
(1) We introduce a novel class of disturbance observers for legged robots, based on measurable variables and incorporating dynamic gains. Unlike the approaches in\cite{krishnamurthy2020dynamic13, song2016adaptive15, song2017time14}, which employ specific time-varying functions to represent dynamic gains, we develop a class of $\mathcal{K}_F$ functions to serve as dynamic gains. These $\mathcal{K}_F$ functions are coupled with comparison functions and feedback into the disturbance observer to achieve predefined convergence, including ultimately uniformly bounded, asymptotic, and exponential convergence, among other types. In the framework of our proposed disturbance observer, engineers have two options for achieving predefined convergence: one is selecting based on the properties of the $\mathcal{K}_F$ functions, and the other is based on the sufficient conditions that need to be met by the comparison functions, or a combination of both.

(2) Compared to existing disturbance observers such as \cite{kim2010disturbance16,kim2015disturbance17}, which require information on the upper bound of the disturbance or its time derivative to achieve asymptotic or exponential convergence of the disturbance estimation error, our proposed observer benefits from the introduction of $\mathcal{K}_F$  functions. The growth pattern and supremum of the $\mathcal{K}_F$  functions can be tailored, enabling asymptotic or exponential convergence without the need for upper bound information of the disturbance or its derivative. This design flexibility enhances the engineering potential of our disturbance observer.

(3) To validate the effectiveness of our proposed observer, we conducted locomotion control simulations and experiments on a legged robot, comparing the scenarios with and without observer compensation. The results demonstrate that the robot equipped with the disturbance observer exhibits superior fault tolerance and enhanced robustness compared to the robot operating without the disturbance observer.

The rest of this study is organized as follows. Section \ref{sec:Notations-and-Problem} presents the preliminaries. Section \ref{sec:main_results} elaborates on the observer design and convergence analysis.  Section \ref{sec:simulation} gives the simulation and experimental results, and Section \ref{sec:Conclusion} concludes the study.

	\section{Preliminaries\label{sec:Notations-and-Problem}}
	\subsection{Notations}
	$\mathbb R$, $\mathbb R_{\geq 0}$, $\mathbb R_{>0}$ and $\mathbb R^n$ denote the set of real numbers, the set of non-negative numbers, the set of positive numbers, and $n$-dimensional Euclidean space, respectively. $t_0$ denotes the initial time.  For a vector $x = [x_1,\cdots,x_n]\t\in \mathbb R^n$, $\|x\|= \sqrt{x_1^2 +\cdots,x_n^2}$ denotes its Euclidean ($\mathcal L_2$) norm,  and $x>0$ (or $x\geq 0$) means $x_i>0$ (or $x_i\geq 0$) for $i=1,\cdots,n$. A continuous function $\alpha:[0,a)\to [0,\infty)$ is said to belong to class $\mathcal K$ if it is strictly increasing and $\alpha(0)=0$. It is said to belong to class $\mathcal K_\infty$ if $a=\infty $ and $\alpha(r) \to \infty $ as $r\to \infty$. 
	\subsection{Problem formulation}
	According to \cite{zhu2023proprioceptive01}, when the robot moves in the environment, its resultant dynamic model can be written as 
	\begin{equation}\label{eq:SysDynamics}
		\bm M (\bm q)\ddot {\bm q} + \bm C(\bm q,\dot {\bm q })\dot {\bm q} + \bm G(\bm q) = \bm S\t \bm \tau + \bm J_{st}\t \bm F_r + \bm d(t)
	\end{equation}
	where $\bm M(\bm q) \in \mathbb R^{(6+n)\times (6+n)}$, $\bm C(\bm q,\dot{\bm q} )\in \mathbb R^{(6+n)\times (6+n)}$ and $\bm G(\bm q) \in \mathbb R^{(6+n)}$ are the inertia matrix, Coriolis matrix and gravity vector, respectively; $\bm S =[\bm 0_{n\times 6} \quad \bm I_n]$ is the actuated part selection matrix; $\bm \tau \in \mathbb R^n$ is the actuation torques; $\bm F_r \in \mathbb R^{3n_c}$ is the ground reaction forces (GRFs) of $i$-th leg; $\bm J_{st}\in \mathbb R^{3n_c\times (6+n)}$ is the Jacobians that transposes the GRFs into the acceleration of the COM and actuated joints; $n$ and $n_c$ represents the number of active degrees of freedom (DoF) and the number of legs in the support state, respectively; $\bm d :[t_0,\infty ) \to \mathbb D$ denotes the external disturbances, where $\mathbb D$ is a compact set belongs to $\mathbb R^{6+n}$. According to \cite{dong2022fixed18,spong2020robot19}, the matrices $\bm M(\bm q)$, $\bm C(\bm q,\dot {\bm q})$ and $\bm G(\bm q)$ satisfy the following properties. 
	
\bpro \label{prop:1}
For any $\bm q,\dot {\bm q} \in \mathbb R^{6+n}$, there exist positive constants $k_{\overline m }$, $k_{\underline m}$, $k_c$ and $k_g$ such that $k_{\underline m }I_{6+n} \leq  \bm M(\bm q)\leq k_{\overline m}I_{6+n}$, $\|\bm C(\bm q,\dot {\bm q})\|\leq k_c \|\dot {\bm q}\|$ and $\|\bm G(\bm q)\|\leq k_g$. 
\epro

\bpro \label{prop:2}
For any $\bm q,\dot {\bm q}\in \mathbb R^{6+n}$, $\dot {\bm M}(\bm q) = \bm C\t (\bm q,\dot {\bm q}) + \bm C(\bm q,\dot {\bm q} )$. 
\epro

	In this paper, it is assumed that $\bm q$, $\dot{\bm q}$,
	$\bm M (\bm q )$, $\bm C(\bm q,\dot{\bm q})$, $\bm G(\bm q)$, $\bm S \t \bm \tau$ and $\bm J_{st}\t \bm F_r$ are available. 
	According to \cite{lu2019adaptive04}, the time-varying external disturbance $\bm d(t)$ satisfies the following assumption. 
	\bass \label{ass:1}
	The external disturbance $\bm d(t)$ is first-order differentiable satisfying 
	\begin{gather*}
			\sup_{t\in [t_0,\infty)}\| {\bm d }(t)\| =  {{\bm d}}_\m  <\infty, \quad
				\sup_{t\in [t_0,\infty)}\|\dot {\bm d }(t)\| =  {\dot {\bm d}}_\m  <\infty
		\end{gather*}
where $ {{\bm d}}_\m$ and ${\dot {\bm d}}_\m$ are unknown finite constants. 
	\eass
	
The following assumption is introduced to ensure the stable locomotion of the legged robot.
\bass \label{ass:2}
The term $\bm S\t \bm \tau + \bm J_{st}\t \bm F_r $ generates bound position $\bm q$ and velocity $\dot {\bm q}$ for $t\geq t_0$, i.e., 
	\begin{gather*}
	\sup_{t\in [t_0,\infty)}\| {\bm q }(t)\| =  {{\bm q}}_\m  <\infty, \quad 
	\sup_{t\in [t_0,\infty)}\|\dot {\bm q }(t)\| =  {\dot {\bm q}}_\m  <\infty
\end{gather*}
where $ {{\bm q}}_\m$ and ${\dot {\bm q}}_\m$ are  finite constants. 
\eass

We give the following definition to characterize a class of time-varying functions. 

	\bdefn [$\mathcal K_F$ Functions] \label{Def:K_T} 
Define 
\begin{equation}
\mathbb{R}_{p}=[\underline b,\overline b) \label{eq:R_p}
\end{equation}
 with
$0\leq \underline b< \overline b$, where $\overline b$ may tend to infinity, i.e., $b=\infty$. 
A continuous differentiable function $\mu(t):[t_0,\infty)\to\mathbb{R}_{p}$
is said to belong to class $\mathcal{K}_{F}$ if it is strictly increasing with respect to $t$ and
\begin{gather}
	\mu(t_0) = \underline b, \quad \lim_{t\to \infty} \mu(t) =\overline b,\label{eq:mu_derivative}\\
	\frac{\mathrm{d}\mu(t)}{\mathrm{d}t}\leq\tilde{b}(\mu(t))^{2},\quad\forall t\geq t_0\label{eq:bound_mu_derivative}
\end{gather}
where $\tilde{b}$ is some positive constant possibly related
to $\underline b$.  \edefn
\brem
There are many functions $\mu(t)$ that belong to class $\mathcal{K}_F$. For instance, consider $\mu(t) = k_1(t-t_0) + k_2$, with $k_1, k_2 > 0$. In this instance, we have $\underline{b} = k_2$, $\overline{b} = \infty$, and $\tilde{b} = k_1/k_2^2$. Another example is $\mu(t) = \exp \left(k(t-t_0)\right)$, where $k > 0$. Here, $\underline{b} = 1$, $\overline{b} = \infty$, and $\tilde{b} = k$. Additionally, we can consider the scenario where $\mathbb{R}_p$ in \eqref{eq:R_p} is a compact set, implying that $\overline{b} < \infty$. For example, $\mu(t) = \frac{k}{1 + k \exp(-\lambda(t-t_0))}$ with $k, \lambda > 0$. Then, $\underline{b} = \frac{k}{1+k}$, $\overline{b} = k$, and $\tilde{b} = \lambda$.
\erem

%
	
\textbf{Objective}: For the unknown external disturbance $\bm d(t)$, 
the objective of this work is to design external disturbance observer $\hat {\bm d}$ such that the  estimate error 
	\begin{equation} \label{eq:tilde_d}
		\tilde {\bm d }(t) = \hat {\bm d}(t) - \bm d(t)
	\end{equation}
	achieves the predefined convergence, i.e., 
\begin{equation} \label{eq:objective}
	\|\tilde {\bm d}(t)\| \leq  (\gamma(\mu(t)))^{-1} \big (\alpha_1 (\|\tilde {\bm d}(t_0)\|) +\alpha_2 ( {\bm d}_\m) +\alpha_3(\dot {\bm d}_\m) \big)
\end{equation}
where $\alpha_1,\alpha_2,\alpha_3, \gamma \in \mathcal K_\infty$, 
	$\tilde {\bm d}(t_0)$ is the initial value of estimate error $\tilde {\bm d}(t)$, and $\mu(t)\in \mathcal K_F$.

\brem 
The predefined convergence of the disturbance estimation  error $\tilde {\bm d} (t)$ depends on the design of $\mu(t)$ and $\gamma(s)$. For instance, if we design $\mu(t) = \exp(k(t-t_0))$ with $k>0$, this $\mu(t)$ function can serve as a base function for designing $\gamma(s)$ to achieve various desired convergence behaviors. For example, setting $\gamma(s) = cs$ with $c>0$ yields exponential convergence. Alternatively, designing $\gamma(s) = \exp(s) - 1$ can achieve super-exponential convergence as defined in \cite{wang2019general20}. 
\erem

\section{Observer Design and Convergence Analysis }\label{sec:main_results}
In this section, we first elaborate the observer design, and then prove the stability and convergence of the proposed observer's dynamics.	\subsection{External Disturbance Observer Design }
The external disturbance observer is designed  as  follows: 
\begin{equation}
\hat {\bm d} (t) = \bm M(\bm q) \hat {\bm D}(t) \label{eq:hat-d}
\end{equation} 
where $\hat {\bm D}$ is specified as 
\begin{equation}
		\hat {\bm D}( t) = \bm \xi (t) + \alpha(\mu )  \dot {\bm  q}.  \label{eq:hat_d}
	\end{equation}
In this context, $\alpha \in \mathcal K_\infty$ is first first-order differentiable,  and 
	$\mu(t) \in \mathcal K_F$. The $\bm \xi$-dynamics is designed as 
	\begin{align}
		\dot {\bm \xi } (t) & = -\dot \alpha(\mu )  \dot {\bm q}  - \alpha(\mu ) \bm{M}^{-1} (\bm q) \big[\bm S\t \bm \tau + \bm J_{st}\t \bm F_r\notag \\
		& \quad  -\bm C(\bm q,\dot {\bm q })\dot {\bm q} - \bm G(\bm q)\big]-\alpha(\mu) \hat {\bm D}.  \label{eq:xi-dynamics}
	\end{align} 
	
	\subsection{Stability and Convergence Analysis}
In \eqref{eq:hat-d}, $\hat {\bm D}(t)$ denotes the estimate of 
	$
	\bm D(t) = \bm M^{-1} (\bm q) \bm d(t) \label{eq:D}
	$. Define the corresponding estimate error as   
	\begin{equation}
		\tilde {\bm D}(t) = \hat {\bm D}(t) -\bm D(t).  \label{eq:tilde_D}
	\end{equation}
By \eqref{eq:SysDynamics}, one has 
\begin{equation} \label{eq:ddot_q}
	\ddot {\bm q}   = \bm M^{-1} (\bm q) \big [\bm S\t \bm \tau + \bm J_{st}\t \bm F_r + \bm d(t) - \bm C(\bm q,\dot {\bm q })\dot {\bm q} - \bm G(\bm q) \big]. 
\end{equation}
	According \eqref{eq:hat_d}, \eqref{eq:xi-dynamics}, \eqref{eq:tilde_D} and \eqref{eq:ddot_q},  taking the time derivative of $\tilde {\bm D} (t)$ yields 
	\begin{align}
		\dot {\tilde {\bm D}}(t) & = \dot {\hat {\bm D}}(t) - \dot {\bm D}(t)\notag \\
		& = \dot {\bm \xi}  + \dot \alpha(\mu ) \dot {\bm q }  + \alpha(\mu ) \ddot {\bm q} -\dot {\bm D} \notag \\
		& = -\dot \alpha(\mu ) \dot {\bm q} -\alpha(\mu ) \bm M^{-1}(\bm q)\big[\bm S\t \bm \tau + \bm J_{st}\t \bm F_r -\bm C(\bm q,\dot {\bm q })\dot q\notag \\
		& \quad  - \bm G(\bm q)\big] -\alpha(\mu) \hat {\bm D}+ \dot \alpha(\mu ) \dot {\bm q }  + \alpha(\mu ) \bm M^{-1} (\bm q)\big[ \bm S\t \bm \tau \notag \\
		& \quad  + \bm J_{st}\t \bm F_r + \bm d - \bm C(\bm q,\dot {\bm q })\dot q - \bm G(\bm q)\big ] - \dot {\bm D}	\notag \\ 
		& = -\alpha (\mu) \hat {\bm D}  + \alpha(\mu ) \bm D - \dot {\bm D}\notag \\ 
		& = - \alpha(\mu ) \tilde {\bm D} - \dot {\bm D} \label{eq:tilde_d-dynamics}
	\end{align}
where $\dot {\bm D}$ can be expressed as 
\begin{equation}\label{eq:D-dynamics}
	\dot {\bm D} = -\bm M^{-2}(\bm q) \dot {\bm M}( \bm q) \bm d + \bm M^{-1}(\bm q) \dot {\bm d}. 
\end{equation}

	We then have the following theorem. 
	\bthm \label{the:1}
	Consider the legged robot described in \eqref{eq:SysDynamics}, which adheres to Properties \ref{prop:1}--\ref{prop:2}. Assume that Assumptions \ref{ass:1}--\ref{ass:2} are satisfied. For the disturbance observer $\hat {\bm d}$ designed in \eqref{eq:hat-d}, with $\hat {\bm D}$ and $\bm \xi$-dynamics are specified in \eqref{eq:hat_d} and \eqref{eq:xi-dynamics}, respectively, achieving the objective in \eqref{eq:objective} is feasible for any $\mu \in \mathcal K_F$, provided that $\alpha(s)$ is selected to satisfy 
		\begin{equation}
	\frac{\mathrm d \alpha(s)}{ \mathrm ds} \leq \frac{1}{2} s^{-2} {\tilde b}^{-1}  \sigma (\alpha(s) )^2 \label{eq:partial_alpha_1} 
	\end{equation}
	for any $s\in \mathbb R_{> 0}$ and $0<\sigma <1$. Specifically, the objective is accomplished  with 
	\begin{equation}
			\begin{gathered}
			\gamma(s) = \alpha(s), \quad \alpha_1(s) = k_{\overline m} k_{\underline m}^{-1} \alpha(\mu(t_0))s  \\
			\alpha_2(s) =  2 k_{\overline m } \tilde \sigma ^{-\frac{1}{2}}  k_{\underline m}^{-2}k_c s,\quad \alpha_3(s) = k_{\overline m } \tilde \sigma ^{-\frac{1}{2}} k_{\underline m} ^{-1}s. 
		\end{gathered} \label{eq:objective_1}
	\end{equation}
where $\tilde \sigma$ is some constant satisfying $0<\tilde \sigma <1$. 
	\ethm 
	\proofnow 
	We consider the Lyapunov function candidate for $\tilde {\bm D}$-dynamics as 
	\begin{equation} \label{eq:V_Lya}
		V(\tilde {\bm D}) = \tilde {\bm D} \t \tilde {\bm D}. 
	\end{equation}
	Then by \eqref{eq:tilde_d-dynamics}, the time derivative of $V(\tilde {\bm D}) $ is 
	\begin{align}
		\dot V(\tilde {\bm D}) & =2   {\tilde {\bm D}}\t 
		\dot {\tilde {\bm D}}  = -2 \alpha(\mu ) \tilde {\bm D} \t \tilde {\bm D} -2 \tilde {\bm D} \t \dot {\bm D}.  \label{eq:dot_V}
	\end{align}
By Properties \ref{prop:1}--\ref{prop:2} and Assumptions \ref{ass:1}--\ref{ass:2}, the $\dot {\bm D}$ in \eqref{eq:D-dynamics} satisfies 
\begin{align}
\| \dot {\bm D} \| & \leq \| \bm M^{-2} (\bm q) \| \|\bm C\t (\bm q ,\dot {\bm q}) + \bm C (\bm q ,\dot {\bm q}) \|  \| \bm d \| \notag \\
& \quad + \| \bm M^{-1} (\bm q) \| \| \dot {\bm d} \|  \notag \\
& \leq 2 k_{\underline m}^{-2}k_c \dot {\bm q} _\m \bm d_m  + k^{-1}_{\underline m} \dot {\bm d}_m := \dot {\bm D}_m  \label{eq: bound_dot_D}
\end{align}
where we note $\dot {\bm D}_\m $ is a finite constant. 
	Upon using Young's inequality and \eqref{eq: bound_dot_D}, one has 
	\begin{align}
		-2 \tilde {\bm D} \t \dot {\bm D}  &\leq 2 \|\tilde {\bm D} \| \dot {\bm D}_\m \leq \alpha(\mu ) \|\tilde {\bm D}\|^2 + (\alpha(\mu ))^{-1} \dot {\bm D}_\m  ^2 . \label{eq:Yound-Term}
	\end{align}
	Substituting \eqref{eq:Yound-Term} into \eqref{eq:dot_V} yields 
	\begin{equation}
		\dot V (\tilde {\bm D}) \leq -\alpha(\mu ) V(\tilde {\bm D}) + (\alpha(\mu ))^{-1} \dot {\bm D}_\m ^2 \label{eq:bound-dot_V}
	\end{equation}
	We use the changing supply function method in \cite[Section 2.5]{chen2015stabilization02} to further explore the convergence of $\tilde {\bm D}(t)$.  Define a new Lyapunov function candidate as 
	\begin{equation}
		U (\tilde {\bm D}, \mu) = k (\alpha(\mu ))^2  V(\tilde {\bm D}) \label{eq:U_Lya}
	\end{equation}
	where $k>0$.  
	Then by \eqref{eq:bound-dot_V}, 
	the time derivative of $U (\tilde {\bm D}, \mu(t))$ satisfies 
	\begin{align}
		&\dot U(\tilde {\bm D }, \mu(t)) \notag \\
		&= k\alpha(\mu) \dot \alpha(\mu )V(\tilde {\bm D}) + k(\alpha(\mu ))^2  \dot V(\tilde  {\bm D}) \notag \\
		& \leq   k(\alpha(\mu ))^2\big( -\alpha(\mu ) V(\tilde {\bm D}) + (\alpha(\mu ))^{-1} \dot {\bm D}_\m ^2 \big) \notag \\
		& \quad + k\alpha(\mu) \dot \alpha(\mu )V(\tilde {\bm D}) \notag \\
	&	= \big(-\alpha(\mu ) + 2\dot\alpha(\mu ) /\alpha(\mu )  \big) U(\tilde {\bm D},\mu(t)) +k  \alpha(\mu )\dot {\bm D}_\m ^2. \label{eq:dot_U}
	\end{align}
For $\alpha(s)$ and $\mu(t)$ satisfying \eqref{eq:partial_alpha_1} and  \eqref{eq:bound_mu_derivative}, respectively,  one has
	\begin{align}
		\frac{\dot \alpha(\mu)}{\alpha(\mu)}  & = \frac{\mathrm d\alpha(\mu)}{\mathrm d \mu} \dot \mu (\alpha(\mu))^{-1} \leq \frac{1}{2}\sigma \alpha(\mu).  \label{eq:partial_alpha}
	\end{align}
	Substituting \eqref{eq:partial_alpha} into \eqref{eq:dot_U}, we have 
	\begin{equation}
		\dot U(\tilde {\bm D},\mu ) \leq -\tilde \sigma \alpha(\mu )U(\tilde {\bm D},\mu )+ k  \alpha(\mu )\dot {\bm D}_\m ^2.\label{eq:dot_U_2}
	\end{equation}
Invoking the comparison lemma in \cite[Lemma 3.4]{2002Nonlinear03} for \eqref{eq:dot_U_2} yields 
	\begin{gather}
		U(\tilde {\bm D}(t), \mu (t)) \leq  \kappa ^{-\tilde \sigma}(\alpha(\mu)) U(\tilde {\bm D}(t_0),\mu(t_0)) \notag \\
		+\int_{t_0}^t \exp \big(-\sint_{\tau }^t \tilde \sigma\alpha(s) \mathrm ds \big) \alpha(\mu(\tau)) k\dot {\bm D}_\m ^2 \mathrm d\tau  \label{eq:Bound_U_1}
	\end{gather}
	where $\kappa^{-\tilde \sigma} (\alpha(\mu))=\exp\left(-\tilde \sigma \sint_{t_{0}}^{t}\alpha(\mu(\tau))\mathrm{d}\tau\right)$. 
	The second term in the right hand side of \eqref{eq:Bound_U_1} satisfies 
	\begin{align}
		&\int_{t_0}^t \exp \big(-\sint_{\tau }^t \tilde \sigma\alpha(s) \mathrm ds \big) \alpha(\mu(\tau)) k\dot {\bm D}_\m ^2 \mathrm d\tau\notag \\
		& = k\dot {\bm D}_\m ^2 \kappa^{-\tilde \sigma }(\alpha(\mu))\int_{t_0}^t \kappa^{\tilde \sigma} (\alpha(\mu(\tau))) \alpha(\mu(\tau)) \mathrm d\tau \notag \\
		& = k\dot {\bm D}_\m ^2 \kappa^{-\tilde \sigma }(\alpha(\mu))\int_{t_0}^t \exp \big( \sint_{t_0}^\tau  \tilde \sigma \alpha(s)\mathrm ds \big) \mathrm d \big(\sint_{t_0}^\tau  \tilde \sigma \alpha(s)\mathrm ds\big)\notag \\
		& = {\tilde \sigma}^{-1} k\dot {\bm D}_\m ^2 \kappa^{-1}(\alpha(\mu))\big(\kappa^{\tilde \sigma}(\alpha(\mu))-1\big ) \notag \\
		& = {\tilde \sigma}^{-1} k\dot {\bm D}_\m ^2\big(1-\kappa^{-\tilde \sigma}(\alpha(\mu)) \big) \label{eq:Sec_Term}
	\end{align}
where  $\kappa^{\tilde \sigma} (\alpha(\mu(\tau)))=\exp\left(\tilde \sigma \sint_{t_{0}}^{\tau}\alpha(\mu(s))\mathrm{d}s\right)$. 
	Substituting \eqref{eq:Sec_Term} into \eqref{eq:Bound_U_1} yields 
	\begin{align}
	&	U(\tilde {\bm D}(t), \mu (t))\notag \\ &\leq \kappa ^{-\tilde \sigma}(\alpha(\mu)) U(\tilde {\bm D}(t_0),\mu(t_0))+ {\tilde \sigma}^{-1}k \dot {\bm D}_\m ^2\big(1-\kappa^{-\tilde \sigma}(\alpha(\mu)) \big) \notag \\
		& \leq U(\tilde {\bm D}(t_0),\mu(t_0)) +  {\tilde \sigma}^{-1} k\dot {\bm D}_\m ^2. 
	\end{align}
By \eqref{eq:V_Lya} and \eqref{eq:U_Lya}, $\tilde {\bm D}(t)$ satisfies 
	\begin{align}
		\|\tilde {\bm D} (t)\| & \leq (\alpha(\mu)) ^{-1} \big( (\alpha(\mu(t_0)))^2  \|\tilde {\bm D}(t_0) \|^2  + \tilde \sigma ^{-1} \dot {\bm D}_\m ^2  \big)^{\frac{1}{2}} \notag \\
		& \leq (\alpha(\mu)) ^{-1} \big( \alpha(\mu(t_0)) \|\tilde {\bm D}(t_0)\| + \tilde \sigma ^{-\frac{1}{2}} \dot {\bm D}_\m  \big)\label{eq:bound_d}
	\end{align}
	where we used the fact $(\sum_{i=1}^n |x_i|)^p \leq \sum_{i=1}^n |x_i|^p$ for $x_i \in \mathbb R,\. i=1,\cdots,n$ and $0<p\leq 1$. 
	
We note $\hat {\bm d} (t) = \bm M(\bm q) \hat {\bm D}(t)$	and $\bm d(t) = \bm M(\bm q) \bm D(t)$. Then the $\tilde {\bm d} (t)$ in \eqref{eq:tilde_d} can be further expressed as 
\[
\tilde {\bm d}(t) = \bm M(\bm q) \big(\hat {\bm D}(t) - \bm D(t)\big) = \bm M(\bm q) \tilde {\bm D}(t).  
\]
Then by Property \ref{prop:1} and \eqref{eq:bound_d}, $\tilde {\bm d}(t)$  satisfies
\begin{align}
\|\tilde {\bm d}(t)\| &\leq k_{\overline m} (\alpha(\mu)) ^{-1} \big( \alpha(\mu(t_0)) \|\tilde {\bm D}(t_0)\| + \tilde \sigma ^{-\frac{1}{2}} \dot {\bm D}_\m  \big) \notag \\
& \leq  (\alpha(\mu)) ^{-1} \big( k_{\overline m} k_{\underline m}^{-1} \alpha(\mu(t_0)) \|\tilde {\bm d}(t_0)\| + k_{\overline m}\tilde \sigma ^{-\frac{1}{2}} \dot {\bm D}_\m  \big).  \notag 
\end{align} 
	Therefore, recall the definition of $\dot {\bm D}_\m$ in \eqref{eq: bound_dot_D} leads to \eqref{eq:objective_1}. 
This completes the proof. 
	\eproof
	\brem 
There are numerous choices for $\alpha(s)$ that satisfy the condition in Theorem \ref{the:1}. For instance, for a given $\mu(t) \in \mathcal{K}_F$ and any $0 < \sigma < 1$, let $\alpha(s) = cs$ where $c \geq 2\tilde{b}/\sigma$. We can verify that $\frac{\mathrm{d} \alpha(s)}{\mathrm{d}s} = c$. Given $c \geq 2\tilde{b}/\sigma$, it follows that $c \leq \frac{1}{2} s^{-2} \tilde{b}^{-1} \sigma (\alpha(s))^2$. Consequently, the condition in Theorem \ref{the:1} is met. Another example is $\alpha(s) = ks \exp(\lambda s)$ where $k \geq 2\tilde{b}/\sigma$ and $\lambda > 0$. We verify that $\frac{\mathrm{d} \alpha(s)}{\mathrm{d}s} = k\exp(\lambda s) + k\lambda s \exp(\lambda s)$. Since $k \geq 2\tilde{b}/\sigma$, we have $\frac{1}{2} \tilde{b}^{-1} \sigma k\exp(\lambda s) \geq 1$ for $s \in \mathbb{R}_{>0}$. Therefore, $\frac{1}{2} \tilde{b} \sigma k\exp(\lambda s) - \lambda s - 1 \geq 0$ for $s \in \mathbb{R}_{>0}$, which implies $k\exp(\lambda s) + k\lambda s \exp(\lambda s) \leq \frac{1}{2} \tilde{b}^{-1} \sigma k^2 s^2 \exp(2\lambda s)$. Hence, the condition in Theorem \ref{the:1} is satisfied.
	\erem

\section{Simulation and Experiment Results}\label{sec:simulation}
In this section, we evaluate the effectiveness of the proposed disturbance observer by comparing the performance of the legged robot with and without disturbance compensation. In \cite{2025arXiv250210761L}, a control framework is introduced to address disturbances in legged systems. We apply the method outlined in \cite{2025arXiv250210761L} to implement disturbance compensation in the legged system. Both simulation and experimental tests are conducted for comparison. The control system is implemented using ROS Noetic in the Gazebo simulator for the simulations. The control architecture and the proposed observer are implemented on a PC (Intel i7-13700KF, 3.4 GHz).  These simulations and experiments are performed on the Unitree A1 quadruped robot. We choose the $\mathcal{K}_{F}$ function described in Definition \ref{Def:K_T} as 
	\begin{gather}
	\mu(t) = \frac{k}{1+k\exp(-\lambda(t - t_0))}\label{eq:kf_function}
	\end{gather}
	with $k,\lambda > 0$.
	\begin{figure}[!ht]
		\centering
		\includegraphics[scale=0.8]{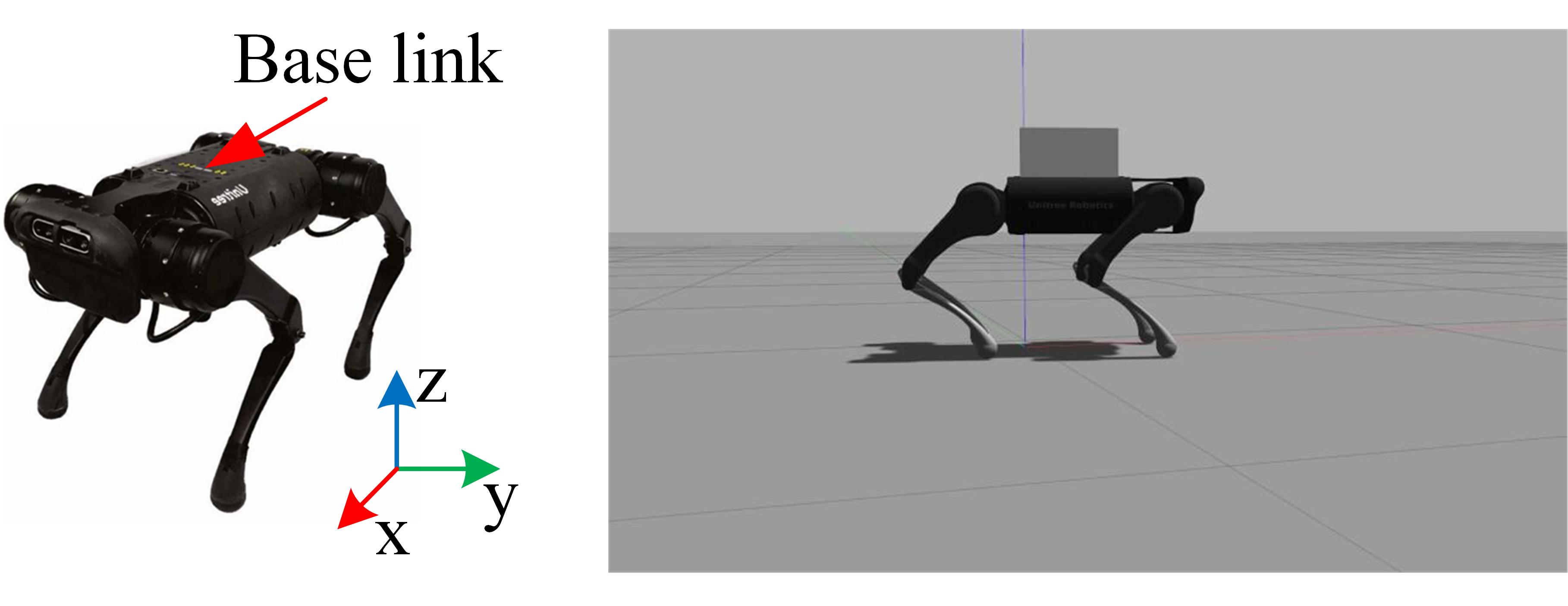}
		\caption{ The Unitree A1 robot (left) and its model in the simulation environment with a 4.5 kg load (right).}
		\label{fig_robot}
	\end{figure}

\subsection{Simulation Comparison}	
\begin{figure}[!ht]
		\centering
		\includegraphics[scale=0.033]{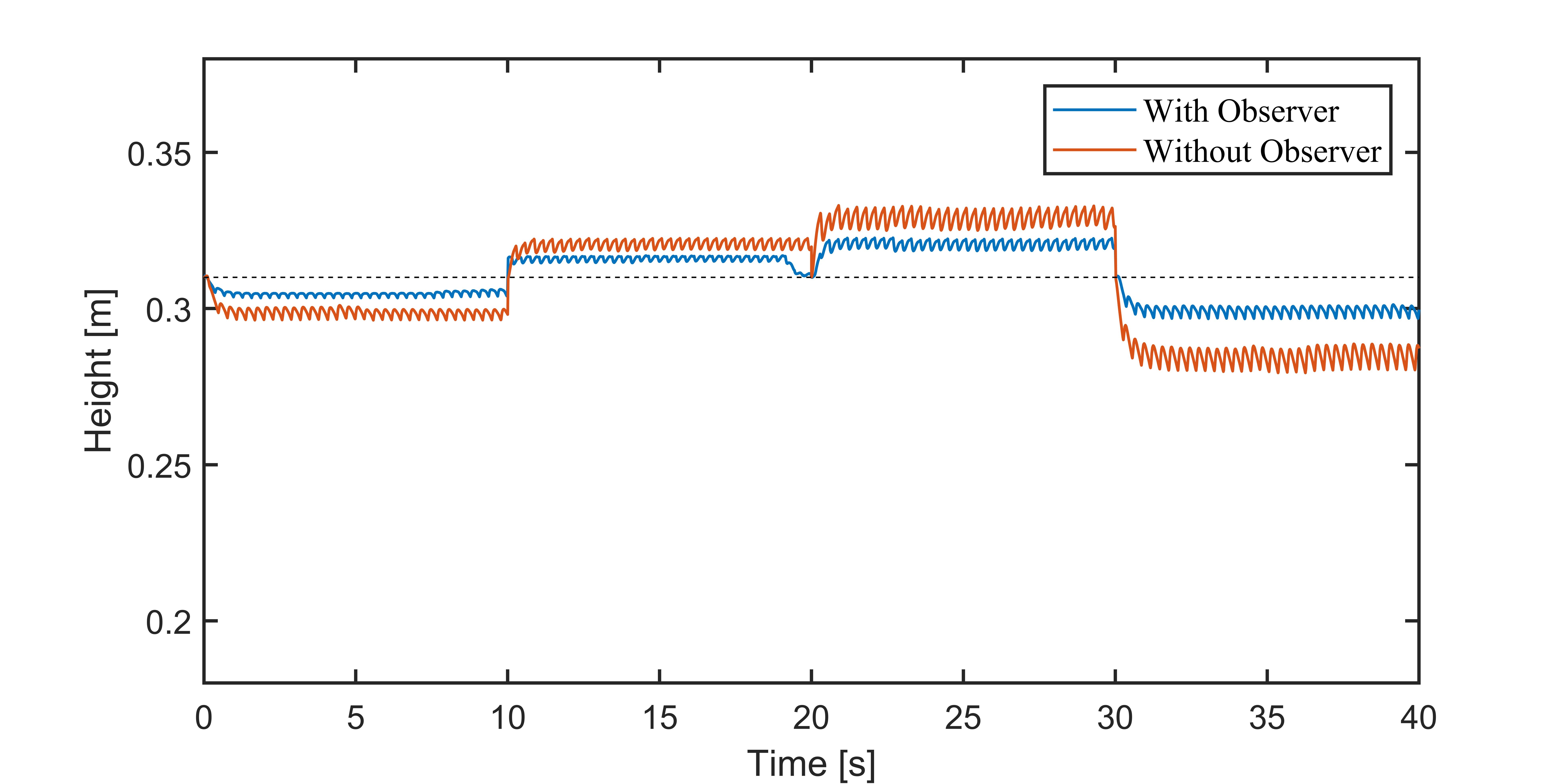}
		\caption{  The height of the base link from the ground over time under the application of different external forces.}
		\label{fig_addforce}
	\end{figure}

The desired height of the base link (see Fig. \ref{fig_robot}) is 0.31 m. Choosing $k = 400$ and $\lambda = 2$ in $\mathcal{K}_{F}$ function (\ref{eq:kf_function}),  and $\alpha(s)= s$ . Figure \ref{fig_addforce} illustrates the height of the base link from the ground over time under the influence of various constant external forces. External forces of $-20$ N, $20$ N, $40$ N, and $-40$ N are applied to the base link along the $z$-axis, each for a duration of 10 seconds. The results demonstrate that the disturbance rejection performance of the legged system with the proposed observer is superior to that of the system without the observer, showing the observer’s ability to effectively estimate disturbances.
	
\begin{figure}[!ht]
		\centering
		\includegraphics[scale=0.55]{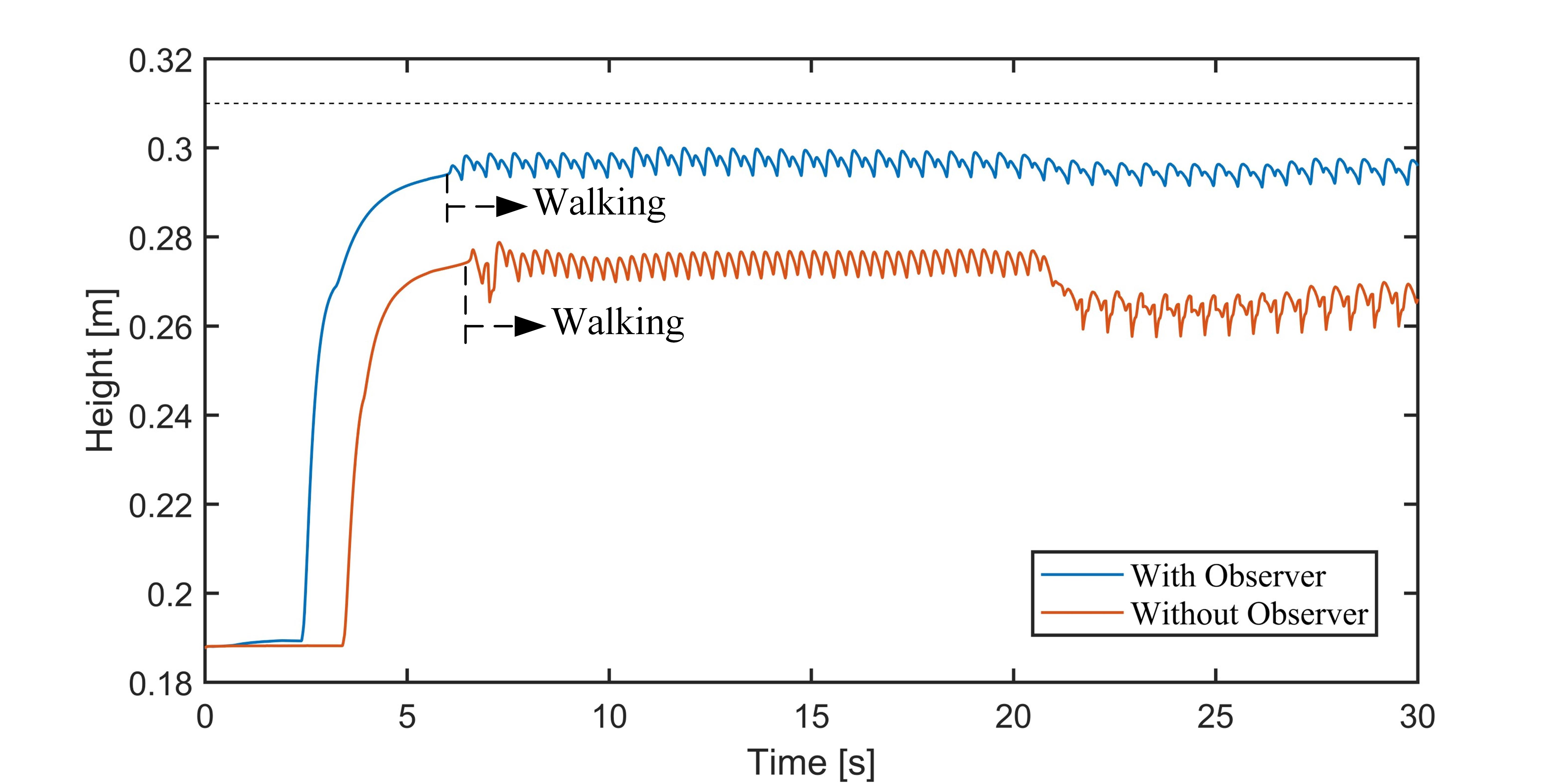}
		\caption{The base link height of the robot with a 4.5 kg load, with and without the observer.}
		\label{fig_load_5kg}
	\end{figure}

	Fig. \ref{fig_load_5kg} shows the height of the base link from the ground over time as the robot walks with a 4.5 kg load at different velocities in the Gazebo simulator. Before 20 seconds, the robot moves at $0$ m/s along the $x$-axis, and after 20 seconds, it moves at $-0.2$ m/s along the $x$-axis. The results demonstrate that the legged system with the proposed observer is more robust than the system without the observer and that the observer is capable of estimating disturbances in the legged system.
	
\subsection{Experiment Comparison}	
\begin{figure}[!ht]
		\centering
		\includegraphics[scale=0.030]{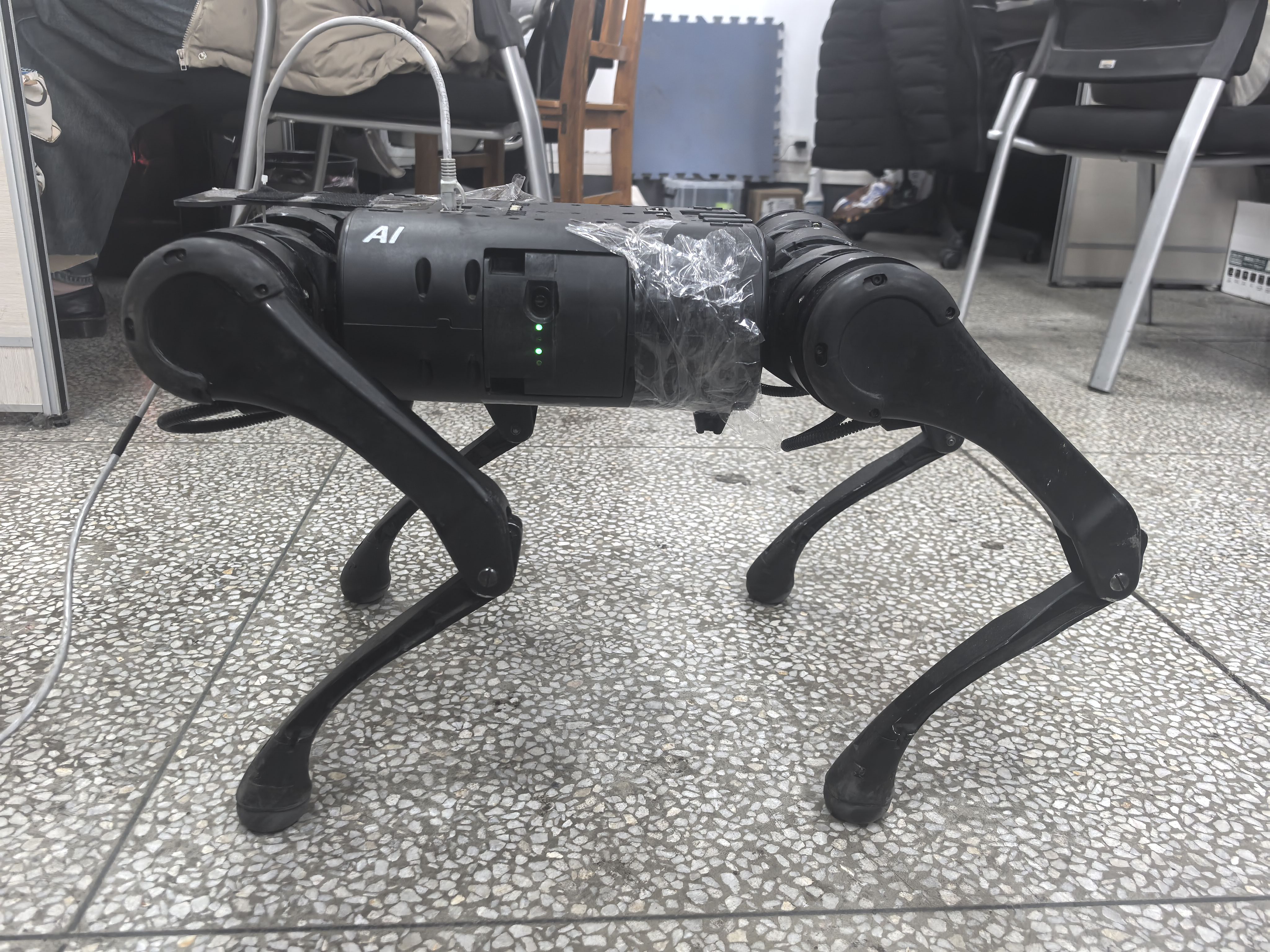}
		\caption{The Unitree A1 robot as the experimental platform.}
		\label{fig_real_robot}
	\end{figure}

\begin{figure}[!ht]
		\centering
		\includegraphics[scale=0.55]{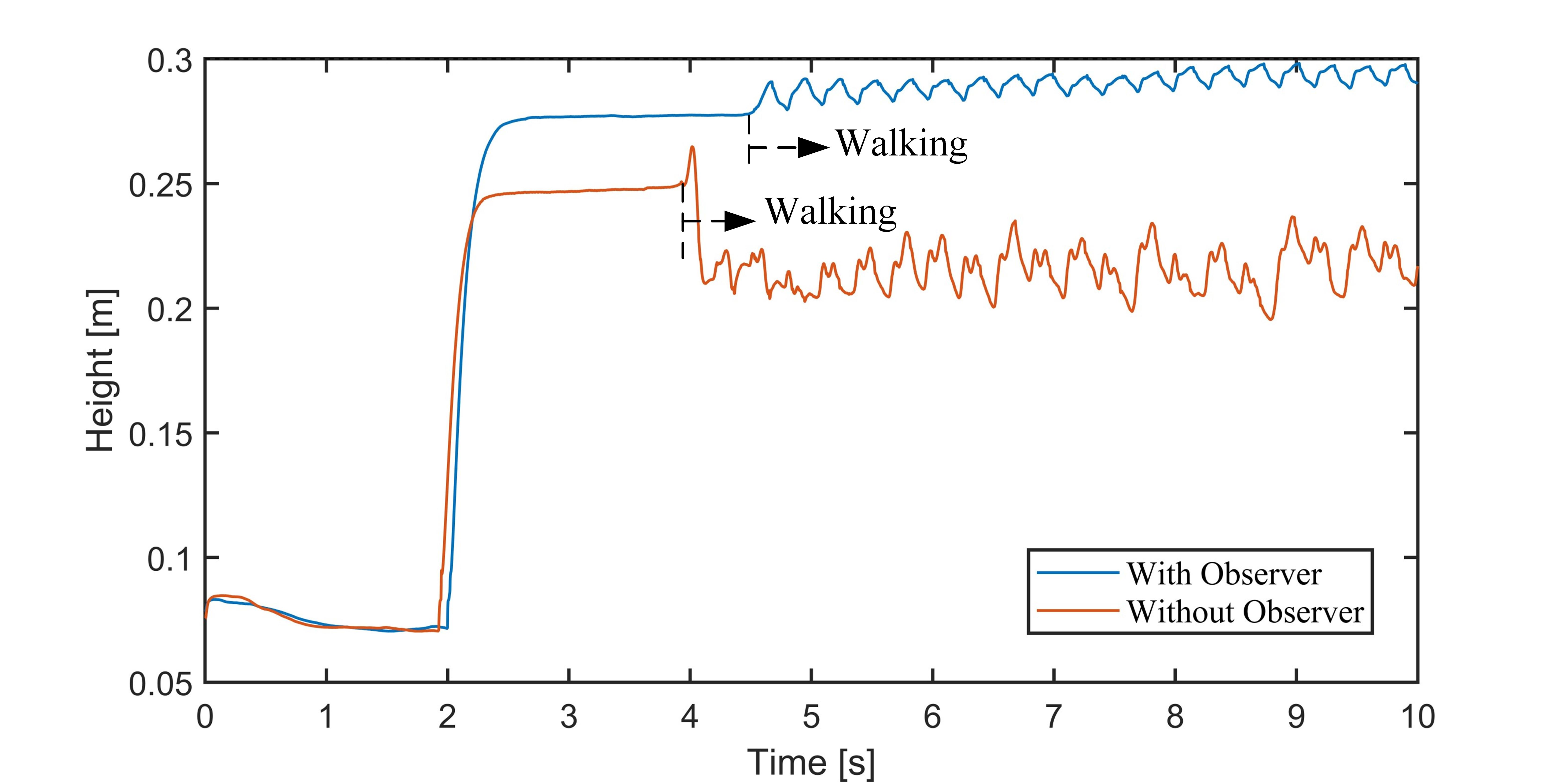}
		\caption{ The height of the base link when the output torque of the robot knee joint is reduced by 50\%, with and without the observer.}
		\label{fig_cutTorque}
	\end{figure}

\begin{figure}[!ht]
		\centering
		\includegraphics[scale=0.033]{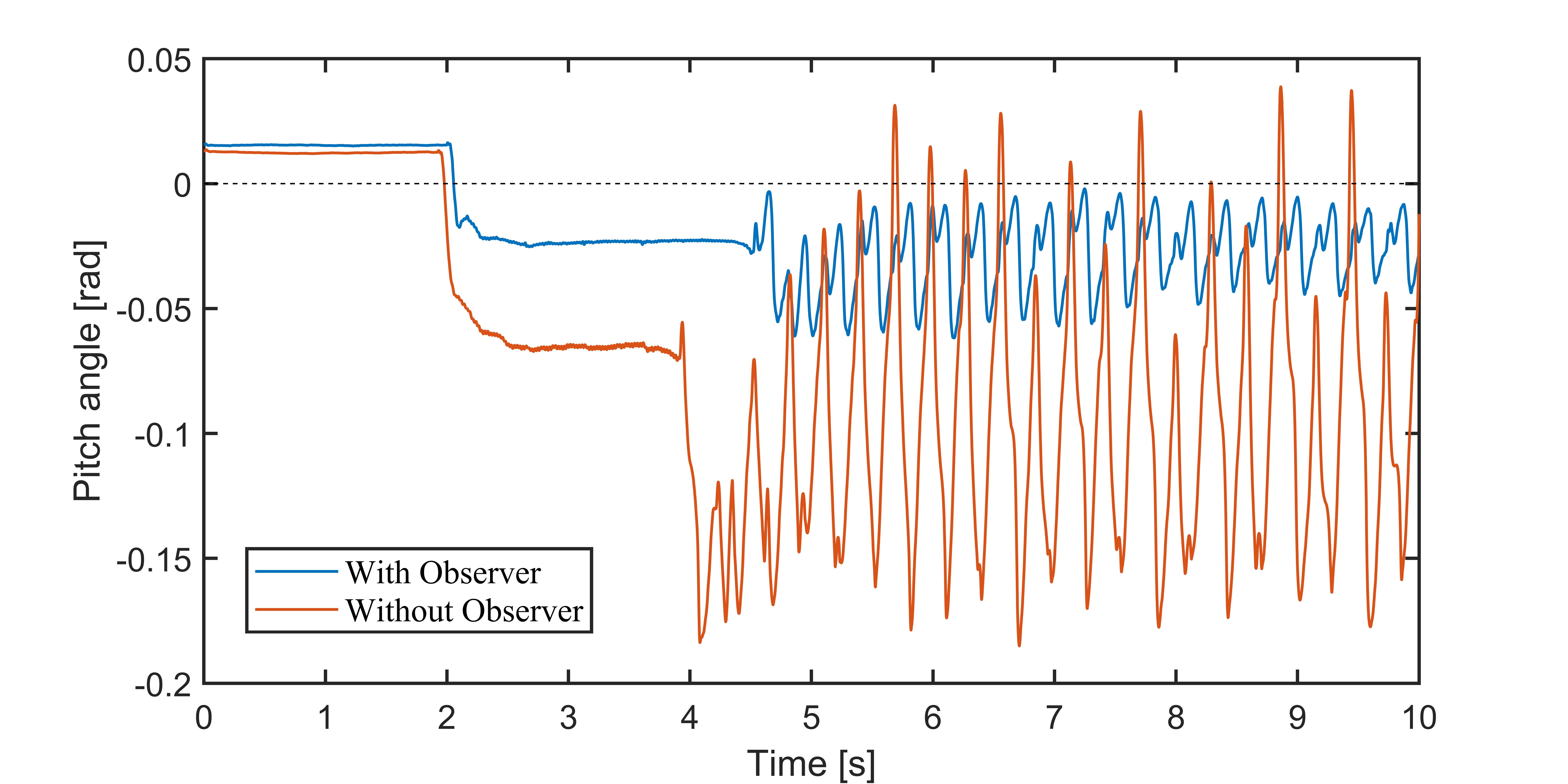}
		\caption{ The pitch angle of the base link when the output torque of the robot knee joint is reduced by 50\%, with and without the observer.}
		\label{fig_pitch}
	\end{figure}

\begin{figure}[!ht]
		\centering
		\includegraphics[scale=0.033]{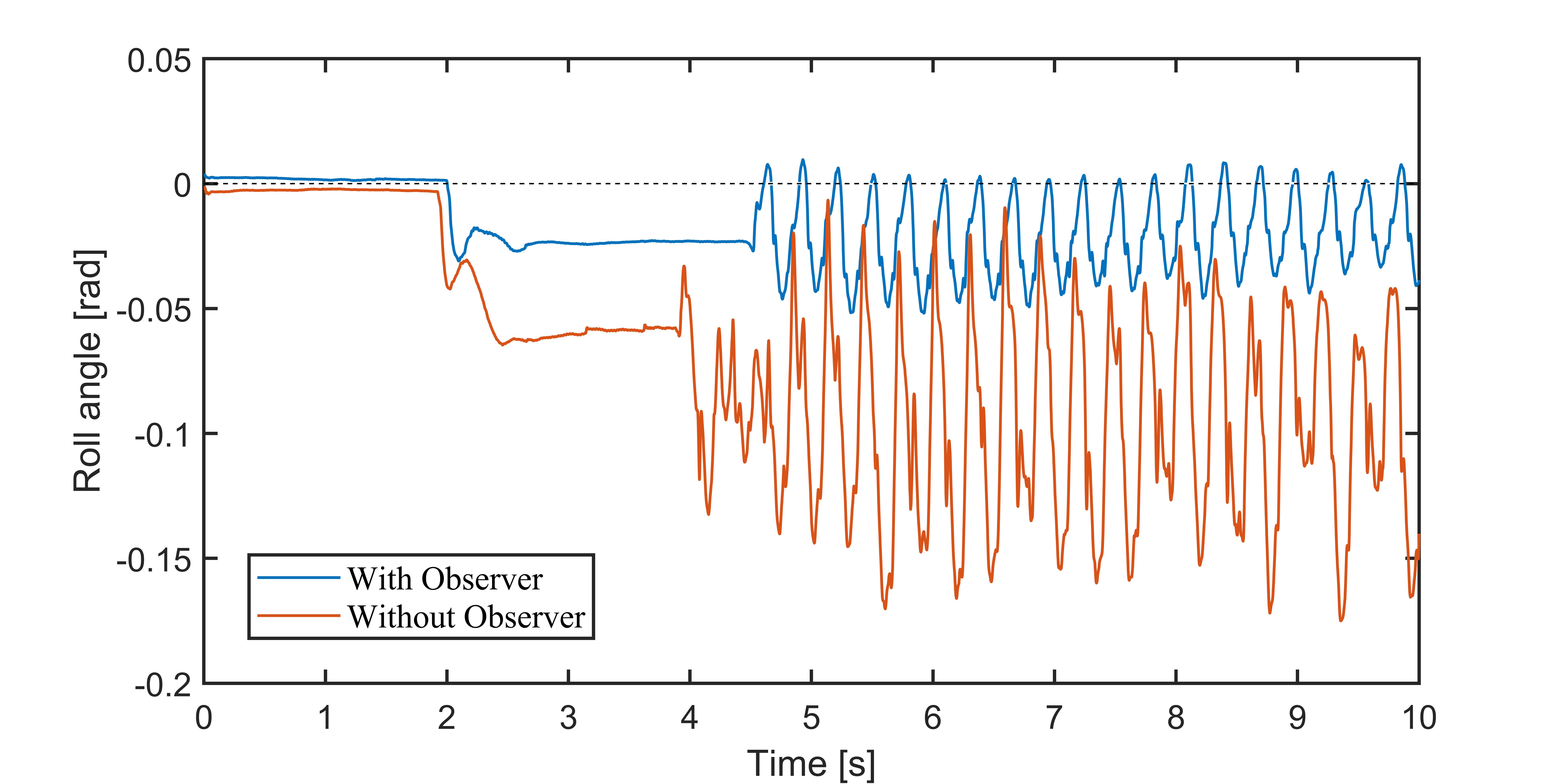}
		\caption{ The roll angle of the base link when the output torque of the robot knee joint is reduced by 50\%, with and without the observer.}
		\label{fig_roll}
	\end{figure}

	In the real-robot experiment, we intentionally reduce the output torque of the robot's knee joint on the right back leg by 50\%. The experiment platform is shown in Fig. \ref{fig_real_robot}. The desired base link height is set to 0.3 m and the desired roll and pitch angles are set to 0 rad. Choosing $k = 200$ and $\lambda = 100$ in $\mathcal{K}_{F}$ function (\ref{eq:kf_function}). The base link height of the robot, as it transitions from standing to stepping in place, is shown in Fig. \ref{fig_cutTorque} both with and without the observer. The roll and pitch angles of the base link are depicted in Figs. \ref{fig_pitch} and \ref{fig_roll}, respectively. For the legged system with the observer, the height, pitch angle, and roll angle trajectories of the base link closely follow the desired trajectories, outperforming the robot without the observer. As shown in Figs. \ref{fig_cutTorque}--\ref{fig_roll}, the robot with the observer demonstrates better fault tolerance and enhanced robustness compared to the robot without the observer.

	\section{Conclusion\label{sec:Conclusion}}
	 This study presents an advanced continuous-time online feedback-based disturbance estimation approach for legged robots, demonstrating broad-ranging convergence including ultimately uniformly bounded, asymptotic, and exponential types. Notably, the method achieves high computational efficiency and precision without the need for explicit upper bounds on disturbances or their derivatives.  Extending this methodology to tackle challenges posed by measurement noise represents a promising avenue for further research, potentially enhancing the robustness and applicability of our findings.

	\bibliographystyle{unsrt}
	\bibliography{ref}

\end{document}